\documentclass[conference]{IEEEtran}
\IEEEoverridecommandlockouts

\usepackage{cite}
\usepackage{amsmath,amssymb,amsfonts}
\usepackage{algorithmic}
\usepackage{graphicx}
\usepackage{textcomp}

\usepackage[table,xcdraw]{xcolor}

\usepackage[colorlinks=true, linkcolor=blue, citecolor=blue, urlcolor=blue]{hyperref}

\usepackage{booktabs}
\usepackage{tabularx}
\usepackage{tcolorbox}
\usepackage{colortbl}
\usepackage{pdflscape}
\usepackage{caption}
\usepackage{url}

\usepackage{tikz}

\def\BibTeX{{\rm B\kern-.05em{\sc i\kern-.025em b}\kern-.08em
    T\kern-.1667em\lower.7ex\hbox{E}\kern-.125emX}}
\begin{document}

\title{Forecast-Driven MPC for Decentralized Multi-Robot Collision Avoidance\\
}

\author{

\IEEEauthorblockN{Hadush Hailu}
\IEEEauthorblockA{\textit{Department of Computer Science} \\
\textit{Maharishi International University}\\
IA, USA \\
hadush.gebrerufael@miu.edu}
\and
\IEEEauthorblockN{Bruk Gebregziabher}
\IEEEauthorblockA{\textit{Electrical Engineering and Computing} \\
\textit{University of Zagreb}\\
Zagreb, Croatia \\
bruk.gebregziabher@fer.hr}
\and
\IEEEauthorblockN{Prudhvi Raj}
\IEEEauthorblockA{\textit{Department of Computer Science} \\
\textit{Eastern Illinois University}\\
Illinois, USA \\
pjallipeta@eiu.edu}
}

\maketitle
\begin{abstract}
The Iterative Forecast Planner (IFP) is a geometric planning approach that offers lightweight computations, scalable, and reactive solutions for multi-robot path planning in decentralized, communication-free settings. However, it struggles in symmetric configurations, where mirrored interactions often lead to collisions and deadlocks. We introduce eIFP-MPC, an optimized and extended version of IFP that improves robustness and path consistency in dense, dynamic environments. The method refines threat prioritization using a time-to-collision heuristic, stabilizes path generation through cost-based via-point selection, and ensures dynamic feasibility by incorporating model predictive control (MPC) into the planning process. These enhancements are tightly integrated into the IFP to preserve its efficiency while improving its adaptability and stability. Extensive simulations across symmetric and high-density scenarios show that eIFP-MPC significantly reduces oscillations, ensures collision-free motion, and improves trajectory efficiency. The results demonstrate that geometric planners can be strengthened through optimization, enabling robust performance at scale in complex multi-agent environments.
\end{abstract}


\section{Introduction}
In recent years, the rapid advancement of autonomous mobile robots has led to growing interest in multi-robot collision avoidance—a key requirement in various applications, including multi-robot search and rescue \cite{b1}, intelligent warehouse systems \cite{b2}, navigation through human crowds \cite{b3}, and autonomous driving \cite{b4}. This task involves enabling each robot to move from its initial location to a designated goal while avoiding collisions with other robots and surrounding obstacles. The complexity of this problem arises from the dynamic interactions among heterogeneous robots and the inherent uncertainty in real-world environments.

To tackle this challenge, coordination strategies generally fall into centralized and decentralized approaches. Centralized frameworks leverage global state information to compute conflict-free paths and have demonstrated success in structured environments \cite{b7, b5, b6, b8}. However, they face scalability issues \cite{b9}, are vulnerable to single points of failure, and lack responsiveness in dynamic settings \cite{b10, b11}.

Decentralized methods, on the other hand, rely on local sensing and onboard computation, enabling greater scalability and robustness in communication-limited scenarios \cite{b10, b19, b15, b17, b13, b16, b12, b18, b14}. While learning-based and optimization techniques offer flexibility and theoretical guarantees, they can be costly or sensitive to domain shifts. Geometric methods, such as velocity cones and Voronoi corridors \cite{b13, b14}, offer a good trade-off between efficiency, reactivity, and provable safety under idealized assumptions, though they may degrade in highly dynamic scenes.

This work introduces a novel planning method, eIFP-MPC, which builds upon the geometric foundation of the Iterative Forecast Planner~\cite{b20, b21}. The method incorporates cost-based via-point selection and integrates the enhanced planner with Model Predictive Control (MPC) to improve collision avoidance and trajectory stability in dense, communication-free multi-robot environments.

\section{Related Works}
Decentralized multi-robot systems aim to achieve coordinated, collision-free navigation using only onboard sensing and local computations, without inter-robot communication. This paradigm ensures scalability and robustness in communication-denied or bandwidth-limited environments but introduces key challenges: maintaining safety guarantees, resolving dynamic conflicts, and preventing deadlocks in constrained spaces. Existing methods fall into four main categories: learning-based, optimization/control-based, sampling-based, and geometric/velocity-space approaches.

\subsection{Reinforcement Learning-Based Approaches}
Reinforcement learning (RL) is widely adopted for learning decentralized multi-robot collision avoidance policies. Key advances include improved reward design, generalization, and observation encoding using attention mechanisms \cite{b22}. Chen et al. \cite{b23} proposed a DDQN model that processes raw LiDAR input on a grid-map, enabling reactive, model-free decision-making. A distributed PPO variant with egocentric occupancy maps was also validated on real robots \cite{b19}. The centralized training with decentralized execution (CTDE) paradigm has enabled joint value learning while preserving decentralized policies \cite{b24}. Frameworks like GA3C-CADRL \cite{b15} scale experience collection using asynchronous agents, and socially-aware DRL models integrate human motion priors for navigating human environments \cite{b17}. RL excels at learning reactive, adaptive behavior from raw sensory inputs in dynamic, unstructured domains. Nonetheless, issues such as high sample inefficiency, lack of formal safety guarantees, and sim-to-real generalization remain open challenges.

\subsection{Optimization and Control-Based Approaches}
Optimization-based methods provide decentralized solutions with formal safety and performance guarantees. RLSS \cite{b13} uses local convex QPs for linear separation, enabling reactive planning with theoretical backing. Zhu et al. \cite{b16} define buffered Voronoi cells for chance-constrained safety under uncertainty. DREAM \cite{b25} incorporates probabilistic modeling of agent intentions for asynchronous decentralized planning, while barrier-certificate MPC \cite{b26} ensures forward invariance using control barrier functions. These methods offer interpretability and enforce hard safety constraints with dynamic feasibility and deadlock avoidance. Still, real-time solvability can be demanding, and overly conservative behaviors may arise in dense or constrained settings without inter-agent coordination.

\subsection{Sampling-Based Planning Approaches}
Sampling-based planners excel in high-dimensional spaces. DEC-LOS-RRT \cite{b12} maintains decentralized safety by restricting sampling to line-of-sight neighbors. Bakker et al. \cite{b18} introduced optimization fabrics—locally reactive trajectories adapting to dynamic obstacles using local sensing. Decentralized PRM variants \cite{b27} improve robustness through adaptive neighborhood resampling. These planners are computationally efficient and probabilistically complete, making them ideal for sparse or exploratory environments. However, they can exhibit limited reactivity and suboptimality, and may behave conservatively when robots plan in isolation.

\subsection{Geometric and Velocity-Space Methods}
Geometric methods model collision avoidance in position or velocity space and are widely valued for their efficiency and scalability in multi-robot systems. Buffered Voronoi Cells~\cite{b14} define safe motion corridors, while randomized separating hyperplanes~\cite{b28} and convexified RVOs~\cite{b29} enable decentralized navigation with safety guarantees. Lu~\textit{et al.}~\cite{b30} integrated barrier functions with MPC to ensure safety and liveness without communication. While effective in swarm and aerial systems, these approaches often assume simple dynamics and limited interaction reasoning, which can lead to oscillations or suboptimal trajectories in dense environments.

To overcome these limitations, we introduce a novel planning method, eIFP-MPC, which extends the geometric collision avoidance of~\cite{b21} through three key contributions: (1) a TCPA-based obstacle ranking mechanism that prioritizes imminent threats during via-point selection, (2) a cost function for via-point evaluation that reduces oscillations and mitigates symmetry-induced deadlocks, and (3) integration of the enhanced strategy with a Model Predictive Control (MPC) framework to ensure smooth and dynamically feasible trajectory tracking. These contributions enable decentralized, communication-free collision avoidance with improved safety, robustness, and planning consistency, as demonstrated through extensive simulation studies.

\section{Methodology}
\subsection{Problem Formulation and Robot Model}

We consider decentralized multi-robot navigation in dynamic environments, where each robot must reach a fixed goal while avoiding static and dynamic obstacles, including other robots. Agents operate independently without communication, relying solely on local state estimation and passive sensing. The objective is to ensure real-time, collision-free navigation with minimal path deviation.

Each robot is modeled as a point-mass circular agent with a safety-augmented radius to account for uncertainty. Motion follows a discrete-time unicycle model:

\begin{equation}
\mathbf{x}_{k+1} = f(\mathbf{x}_k, \mathbf{u}_k) = 
\begin{bmatrix}
x_k + v_k \cos(\theta_k) \Delta t \\
y_k + v_k \sin(\theta_k) \Delta t \\
\theta_k + \omega_k \Delta t
\end{bmatrix}
\end{equation}

where $\mathbf{x}_k = [x_k, y_k, \theta_k]^\top \in \mathbb{R}^2 \times \mathbb{S}^1$ denotes position and heading, and $\mathbf{u}_k = [v_k, \omega_k]^\top$ are the linear and angular control inputs.

Control inputs are subject to bounds:
\begin{equation}
v_k \in [v_{\min}, v_{\max}], \quad \omega_k \in [\omega_{\min}, \omega_{\max}]
\end{equation}

\subsection{Collision Prediction Using TCPA and DCPA}

To assess collision risk with dynamic obstacles, we compute the \textit{Time-to-Closest-Point-of-Approach (TCPA)} and \textit{Distance-at-Closest-Point-of-Approach (DCPA)}, assuming constant velocities.

Let $\mathbf{p}_r, \mathbf{p}_o \in \mathbb{R}^2$ be the robot and obstacle positions, with velocities $\mathbf{v}_r, \mathbf{v}_o \in \mathbb{R}^2$. Define:
\begin{equation}
\begin{aligned}
\mathbf{r} &= \mathbf{p}_o - \mathbf{p}_r, \\
\mathbf{v}_{\text{rel}} &= \mathbf{v}_o - \mathbf{v}_r
\end{aligned}
\end{equation}

TCPA is given by:
\begin{equation}
\text{TCPA} = - \frac{\mathbf{r}^\top \mathbf{v}_{\text{rel}}}{\|\mathbf{v}_{\text{rel}}\|^2}
\label{eq:tcpa}
\end{equation}

DCPA is the projected distance at time TCPA:
\begin{equation}
\text{DCPA} = \left\| (\mathbf{p}_r + \text{TCPA} \cdot \mathbf{v}_r) - (\mathbf{p}_o + \text{TCPA} \cdot \mathbf{v}_o) \right\|
\label{eq:dcpa}
\end{equation}

If $\|\mathbf{v}_{\text{rel}}\|^2 = 0$, then $\text{TCPA} = 0$ and $\text{DCPA} = \|\mathbf{r}\|$.

A collision threat is declared if:
\begin{equation}
\text{DCPA} \leq D_{\text{thresh}}, \quad 0 < \text{TCPA} < T_{\text{max}}
\label{eq:collision_condition}
\end{equation}
where $D_{\text{thresh}} = 5r_{\text{robot}} + d_{\text{safety}}$ and $T_{\text{max}}$ is a time horizon.

Threats are prioritized by sorting TCPA:
\begin{equation}
\text{ThreatRank} = \text{argsort} \left( \{ \text{TCPA}_i \mid \text{collision}_i = \text{True} \} \right)
\end{equation}

This ranking guides the robot to first resolve the most imminent collisions, enhancing responsiveness in dense scenarios.

\begin{figure*}[htbp]
    \centering
    \includegraphics[width=1.0\textwidth]{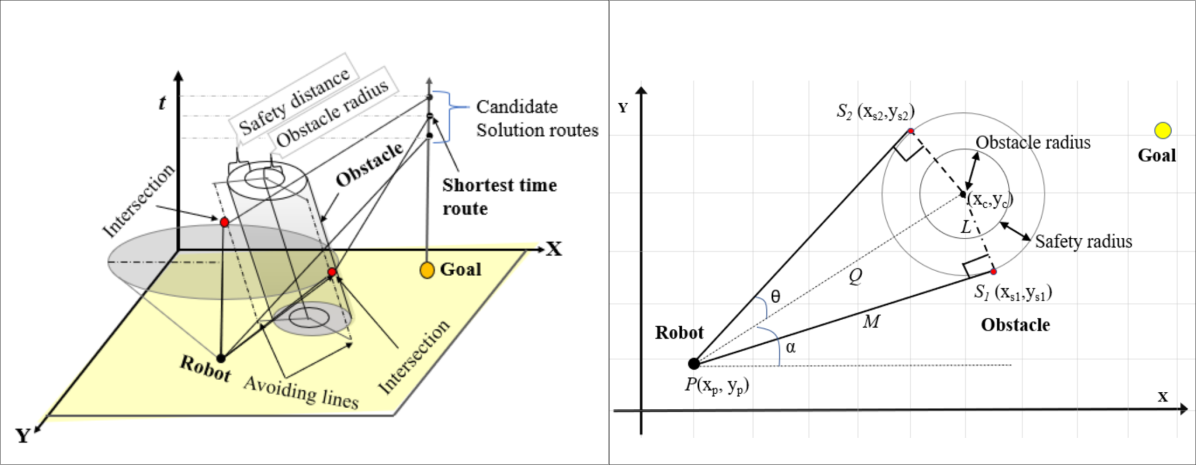}
    \captionsetup{font=footnotesize, justification=raggedright, labelsep=period}
    \caption{Proposed velocity-space collision avoidance method. {Left}: Spatio-temporal configuration space showing the robot’s reachable velocity cone intersecting with tangent-defined avoidance planes. Candidate via-points lie at these intersections, with the shortest-time point selected. {Right}: Geometric construction of tangents from robot position $P(x_p, y_p)$ to an inflated obstacle (radius $L$), forming avoidance directions $S_1$ and $S_2$.}
    \label{fig:Config_Space}
\end{figure*}

\subsection{Geometric-Based Via-Point Planning with Velocity-Space Filtering}

To enable collision-free motion, we generate via-points from tangents to inflated obstacle boundaries and filter them through the robot’s velocity-space cone (Fig.~\ref{fig:Config_Space}). Each obstacle is inflated by a safety margin to define a conservative avoidance boundary.

Let the robot be at $P(x_p, y_p)$ and the obstacle center at $(x_c, y_c)$. The inflated radius and inter-agent distance are:
\begin{equation}
\begin{aligned}
L &= r_{\text{robot}} + r_{\text{obstacle}} + d_{\text{safety}},
Q &= \|P - (x_c, y_c)\|
\end{aligned}
\end{equation}

The heading angle and angular offset are:
\begin{equation}
\begin{aligned}
\alpha &= \tan^{-1}\left(\frac{y_c - y_p}{x_c - x_p}\right), 
\theta &= \sin^{-1}\left( \frac{L}{Q} \right)
\end{aligned}
\end{equation}

The length from robot to tangent contact is $M = \sqrt{Q^2 - L^2}$, and the tangent points are:
\begin{equation}
(x_{s1}, y_{s1})=\left( x_p + M\cos(\alpha - \theta),\ y_p 
 + M\sin(\alpha - \theta) \right)
\end{equation}

\begin{equation}
(x_{s2}, y_{s2}) = \left( x_p + M\cos(\alpha + \theta),\ y_p + M\sin(\alpha + \theta) \right)
\end{equation}

These tangents define avoidance directions aligned with obstacle motion and serve as candidate via-points, later filtered by velocity-cone feasibility and time-to-collision criteria.

\subsubsection{Velocity-Space Cone and Line Intersection}

Assuming constant velocities, potential collisions occur where the motion paths of the robot and obstacle intersect in configuration space $(x, y, t)$. Each path is represented parametrically as:
\begin{equation}
\begin{bmatrix}
x \\ y \\ z
\end{bmatrix}
=
\begin{bmatrix}
p_x \\ p_y \\ p_z
\end{bmatrix}
+
\lambda
\begin{bmatrix}
v_x \\ v_y \\ v_z
\end{bmatrix}
\label{eq:parametric}
\end{equation}
where $\lambda > 0$ is a progression parameter, $\mathbf{p}$ is the starting point, and $\mathbf{v}$ is the direction vector.

Feasible motions are constrained by a velocity-space cone rooted at $(x_p, y_p)$ and aligned with the robot's heading. Its surface is defined as:
\begin{equation}
(x - x_p)^2 + (y - y_p)^2 = r^2 z^2
\label{eq:cone}
\end{equation}
where $r$ is the cone aperture and $z$ denotes time progression.

Substituting Eq.~\eqref{eq:parametric} into Eq.~\eqref{eq:cone} yields a quadratic in $\lambda$:
\begin{equation}
(p_x + \lambda v_x - x_p)^2 + (p_y + \lambda v_y - y_p)^2 = r^2 (p_z + \lambda v_z)^2
\label{eq:intersection}
\end{equation}

Real, positive roots of \eqref{eq:intersection} indicate intersections with the cone. The smallest $\lambda > 0$ gives the earliest dynamically feasible via-point; candidates without valid roots are discarded.

\begin{figure*}[htbp]
    \centering
    \includegraphics[width=1.0\textwidth]{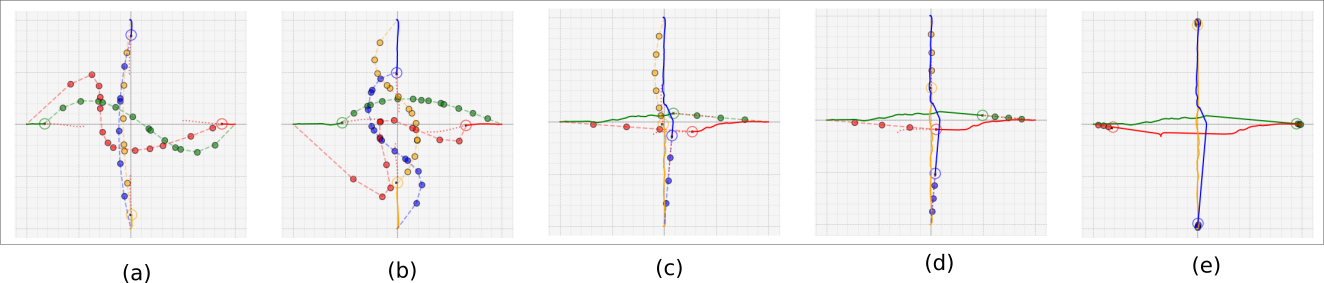}
    \captionsetup{font=footnotesize, justification=raggedright, labelsep=period}
    \caption{Temporal evolution of global plans in a single four-robot scenario using eIFP-MPC. Snapshots (a)--(e) show how each robot’s trajectory adapts over time in response to others’ predicted motions. Early stages exhibit slight plan shifts due to uncertainty, while later frames demonstrate converged, coordinated trajectories.}
    \label{fig:eifpc_mpc}
\end{figure*}

\subsubsection{Cost-Based Via-Point Evaluation and Selection}

To improve stability in symmetric scenarios, via-point selection is framed as a cost-minimization problem over the candidate set $\mathcal{V} = \{\mathbf{v}_i\}$. Each candidate $\mathbf{v}_i \in \mathbb{R}^2$ is evaluated using a weighted cost function incorporating geometric, predictive, and interaction-aware terms:
\begin{equation}
J(\mathbf{v}_i) = w_g \cdot \ell_i + w_d \cdot \sum_{j} \frac{w_j}{d_{ij}} + w_a \cdot \sum_{j} w_j \cdot \phi_{ij}
\label{eq:via_cost}
\end{equation}
where $\ell_i$ is the normalized path length, $d_{ij}$ the perpendicular distance from $\mathbf{v}_i$ to robot $j$'s predicted path, $\phi_{ij}$ the angular deviation from robot $j$'s motion, and $w_j$ a trajectory linearity score. 


Path length is normalized by:
\begin{equation}
\ell_i = \frac{\|\mathbf{p}_r - \mathbf{v}_i\| + \|\mathbf{v}_i - \mathbf{g}\|}{\|\mathbf{p}_r - \mathbf{g}\| + \epsilon}
\label{eq:normalized_length}
\end{equation}
with $\mathbf{p}_r$ as the current position, $\mathbf{g}$ as the goal, and $\epsilon$ a small constant.

Each ghost path is defined by $\{\mathbf{s}_j, \mathbf{g}_j^{\text{ghost}}\}$. Angular misalignment is computed as:
\begin{equation}
\phi_{ij} = 1 - \frac{(\mathbf{v}_i - \mathbf{p}_r)^\top (\mathbf{g}_j^{\text{ghost}} - \mathbf{s}_j)}{\|\mathbf{v}_i - \mathbf{p}_r\| \cdot \|\mathbf{g}_j^{\text{ghost}} - \mathbf{s}_j\|}
\label{eq:angle_penalty}
\end{equation}

Candidates too close to ghost paths are discarded. The selected via-point minimizes the total cost:
\begin{equation}
\mathbf{v}^\star = \arg\min_{\mathbf{v}_i \in \mathcal{V}} J(\mathbf{v}_i)
\end{equation}

This strategy balances goal proximity, motion alignment, and social compliance to ensure robust, conflict-aware navigation.

\subsection{Trajectory Smoothing and MPC Tracking}

After selecting a via-point, the robot generates a feasible trajectory by smoothing discrete waypoints and tracking it using model predictive control (MPC).

Let the waypoint set be $\{ \mathbf{p}_i = (x_i, y_i) \}_{i=0}^{n}$. A continuous path is obtained by fitting cubic splines $S_x(t), S_y(t)$ to $x$ and $y$ coordinates, with $t \in [0, 1]$ as a normalized parameter:
\begin{equation}
\mathbf{p}(t) = 
\begin{bmatrix}
x(t) \\
y(t)
\end{bmatrix}
=
\begin{bmatrix}
S_x(t) \\
S_y(t)
\end{bmatrix}
\end{equation}

This enables continuous evaluation of positions and derivatives for tracking.

The robot follows a discrete-time unicycle model. Given state $\mathbf{x}_k = [x_k, y_k, \theta_k]^\top$ and control $\mathbf{u}_k = [v_k, \omega_k]^\top$, the dynamics are:
\begin{equation}
\mathbf{x}_{k+1} =
\begin{bmatrix}
x_k + v_k \cos(\theta_k) \cdot \Delta t \\
y_k + v_k \sin(\theta_k) \cdot \Delta t \\
\theta_k + \omega_k \cdot \Delta t
\end{bmatrix}
\end{equation}

MPC minimizes tracking error and control effort over a horizon $N$:
\begin{equation}
J = \sum_{k=0}^{N-1} \left( \| \mathbf{x}_k - \mathbf{x}^{\text{ref}}_k \|_Q^2 + \| \mathbf{u}_k \|_R^2 \right)
\end{equation}

where $\mathbf{x}^{\text{ref}}_k$ is the spline reference, and $Q, R \succ 0$ are weighting matrices.

The controller enforces dynamics and bounded inputs:
\begin{equation}
v_{\min} \leq v_k \leq v_{\max}, \quad \omega_{\min} \leq \omega_k \leq \omega_{\max}
\end{equation}

Predicted obstacle trajectories can be incorporated as inequality constraints to ensure safe tracking in dynamic environments.

\section{Experiments and Discussions}
\begin{figure*}[htbp]
    \centering
    \includegraphics[width=1.0\textwidth]{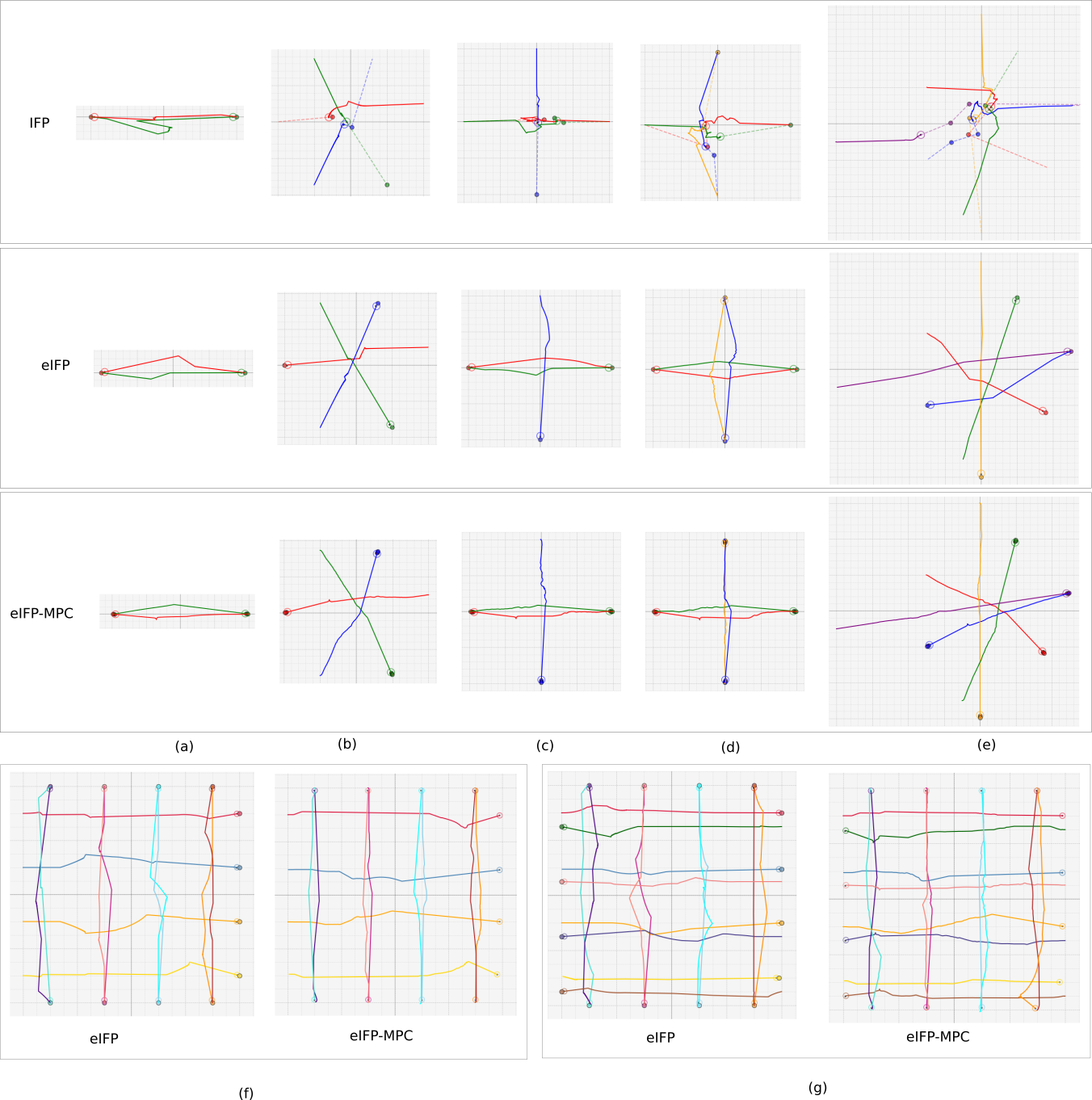}
    \captionsetup{font=footnotesize, justification=raggedright, labelsep=period}
    \caption{Overview of experimental setups used to evaluate the proposed planning methods. Setups (a)--(g) vary in robot count, spatial arrangement, and interaction complexity—from simple pairwise avoidance in (a)--(c) to dense, high-interaction scenarios in (d)--(g). These configurations form the basis for all quantitative and qualitative evaluations throughout the study.}
    \label{fig:experiments}
\end{figure*}

\begin{table*}[ht]
    \centering
    \captionsetup{font=footnotesize, justification=centering, labelsep=period}
    \caption{Quantitative comparison of planning methods across setups (a)--(g). Metrics include average number of oscillations, average time to goal (s), average path length (cm), and average planning time per iteration (ms). The proposed eIFP-MPC consistently reduces oscillations and improves path consistency, especially in high-density scenarios. IFP results are omitted since planner fails to converge due to compounded oscillations and unresolved interactions.}
    \renewcommand{\arraystretch}{1.2}
    \label{tab:eifp_mpc_grouped_flat}
    \begin{tabularx}{\textwidth}{|c|c|*{4}{>{\centering\arraybackslash}X}|}
        \hline
        \rowcolor{red!20}
        \multicolumn{2}{|c|}{\textbf{Approach $\downarrow$}} & \multicolumn{4}{c|}{\textbf{Metrics $\downarrow$}} \\
        \cline{1-2} \cline{3-6}
        \textbf{Control strategy} & \textbf{Number of robots} & \textbf{Average planning time (ms)} & \textbf{Average oscillations} & \textbf{Average path length (cm)} & \textbf{Average time to goal (s)} \\
        \hline
        eIFP      & 2  & 6.711   & 0.0     & 39.85   & 26.567 \\
        eIFP-MPC  & 2  & 54.286  & 0.0     & 37.609  & 29.202 \\
        \hline
        eIFP      & 3  & 8.447   & 12.0    & 39.633  & 30.845 \\
        eIFP-MPC  & 3  & 66.587  & 23.667  & 39.034  & 31.620 \\
        eIFP      & 3  & 13.137  & 85.333  & 41.867  & 27.912 \\
        eIFP-MPC  & 3  & 66.806  & 12.0    & 39.241  & 34.343 \\
        \hline
        eIFP      & 4  & 13.127  & 5.0     & 39.947  & 26.634 \\
        eIFP-MPC  & 4  & 74.733  & 12.75   & 36.852  & 37.353 \\
        \hline
        eIFP      & 5  & 24.529  & 48.0    & 51.664  & 40.469 \\
        eIFP-MPC  & 5  & 79.906  & 11.6    & 48.15   & 65.642 \\
        \hline
        eIFP      & 12 & 69.922  & 86.083  & 81.77   & 90.656 \\
        eIFP-MPC  & 12 & 88.619  & 9.667   & 80.107  & 197.496 \\
        \hline
        eIFP      & 16 & 44.784  & 25.0    & 80.789  & 336.446 \\
        eIFP-MPC  & 16 & 67.722  & 14.25   & 78.698  & 502.260 \\
        \hline
    \end{tabularx}
\end{table*}

\begin{figure*}[htbp]
    \centering
    \includegraphics[width=1.0\textwidth]{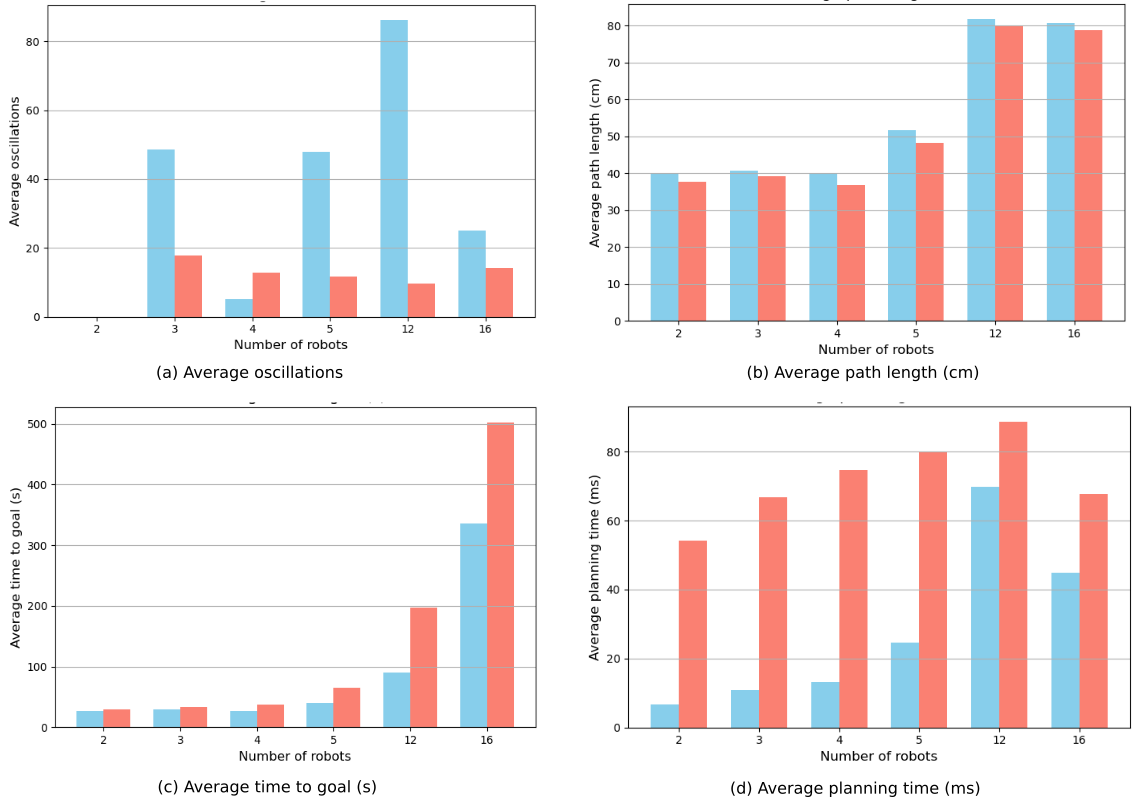}
    \captionsetup{font=footnotesize, justification=raggedright, labelsep=period}
    \caption{Bar plot comparison of planning metrics across setups (a)--(g) for different methods. Each subplot illustrates one metric—average oscillations, average time to goal (s), average path length (cm), or average planning time (ms). The proposed eIFP-MPC achieves lower oscillation rates and smoother paths, with slightly higher planning times, highlighting its stability-performance trade-off.}
    \label{fig:plots}
\end{figure*}

\subsection{Experimental Setup}
To assess the performance of our proposed enhancements to the Iterative Forecast Planner (IFP), we conducted structured simulations in multi-robot navigation scenarios specifically designed to induce oscillatory behaviors—particularly those arising from symmetric configurations and dense interactions.

The experiments were carried out in a two-dimensional planar environment using a custom simulation framework implemented in Python. The simulation used deterministic control loops and fixed random seeds to ensure repeatability. Each robot operated independently in a decentralized manner, with no inter-agent communication or centralized coordination. All robots followed identical behavioral policies and were assigned a velocity of 1.1 grid units per simulation step—interpreted, for convenience, as 1.1\,cm/s.

We defined seven representative scenarios:

\begin{itemize}
    \item {Setup A}: Two robots mirroring each other across a horizontal axis, modeling a minimal symmetry-induced conflict, as shown in Fig.~\ref{fig:experiments}(a).
    
    \item {Setup B}: Three robots forming a T-shaped interaction—one approaching from the top and two from opposite lateral sides (Fig.~\ref{fig:experiments}(b)).

    \item {Setup C}: Three robots arranged at 120° angles, creating radial symmetry to test equal-priority avoidance (Fig.~\ref{fig:experiments}(c)).

    \item {Setup D}: Four robots in a cross configuration, mirrored across both horizontal and vertical axes, as depicted in Fig.~\ref{fig:experiments}(d).

    \item {Setup E}: Five robots randomly distributed with goals converging at the center, inducing a high-density interaction zone (Fig.~\ref{fig:experiments}(e)).

    \item {Setup F}: Twelve robots divided into three groups—four moving left to right, four top to bottom, and four bottom to top—generating structured multi-directional interactions (Fig.~\ref{fig:experiments}(f)).

    \item {Setup G}: Sixteen robots arranged into four directional cohorts, each moving from one boundary to the opposite side, creating dense cross-flow traffic (Fig.~\ref{fig:experiments}(g)).
\end{itemize}

These scenarios were designed to progressively increase geometric symmetry, robot density, and interaction complexity.

\subsection{Evaluation Metrics}

To objectively assess performance across diverse robot configurations and planning strategies, we evaluated four primary metrics. The average planning time (ms) quantifies the mean computational time required for each planning cycle of the control strategy, with lower values indicating higher planning efficiency. The average oscillations metric counts abrupt direction changes or reversals during motion, serving as a proxy for instability and indecisiveness. The average path length (cm) represents the total trajectory length from start to goal, where shorter paths suggest more direct and efficient planning. Finally, the average time to goal (s) captures the duration from the robot's initial position to goal arrival. All metrics were computed by averaging results over 10 independent simulation runs per setup to ensure statistical reliability and account for minor variability.

\subsection{Comparative Evaluation of Planning Methods}

The baseline IFP \cite{b21,b20} algorithm suffers significantly in synchronous multi-robot scenarios, often leading to oscillations, collisions, and inefficiencies. In contrast, both proposed methods—eIFP and eIFP-MPC—demonstrate notable improvements across all setups, with eIFP-MPC providing the most consistent and dynamically feasible behavior. These advantages are evident across the scenario visualizations (Fig.~\ref{fig:experiments}), metric plots (Fig.~\ref{fig:eifpc_mpc}), and quantitative data (Table~\ref{tab:eifp_mpc_grouped_flat}).

In smaller teams (Setups A–C), the differences are subtle. For two robots (Setup A), both eIFP and eIFP-MPC achieve zero oscillations, but eIFP-MPC yields a slightly shorter path (37.61\,cm vs. 39.85\,cm) at the cost of greater planning time (54.29\,ms vs. 6.71\,ms). As robot count increases to three, eIFP begins to suffer from instability—with average oscillations reaching 85.33—while eIFP-MPC keeps this metric below 24. Planning time for eIFP-MPC remains around 66\,ms, showing reliable scalability, whereas eIFP, though faster ($\sim$13\,ms), becomes erratic in symmetry-heavy layouts (e.g., Fig.~\ref{fig:experiments}(b)--(c)).

In more complex interactions (Setups D and E), eIFP’s performance degrades under pressure. For five robots (Setup E), it incurs 48 oscillations and a path length of 51.66\,cm. eIFP-MPC, however, cuts oscillations down to 11.6 and yields a more efficient trajectory (48.15\,cm), albeit with longer goal times (65.64\,s vs. 40.47\,s) due to kinematic enforcement. These improvements are visually clear in Fig.~\ref{fig:experiments}(e), where eIFP paths have long curves, while eIFP-MPC remains almost straight line.

In Setup F (12 robots), eIFP exhibits considerable instability, with an average of 86.08 oscillations and a time-to-goal of 90.66\,s. eIFP-MPC significantly reduces oscillations to 9.67 and marginally shortens the average path length from 81.77\,cm to 80.11\,cm. However, this improvement in stability comes at the expense of a longer execution time (197.50\,s), primarily due to the computational overhead and kinematic feasibility constraints imposed by the MPC controller. In Setup G (16 robots), a similar pattern is observed: eIFP yields 25.0 oscillations and a goal-reaching time of 336.45\,s, whereas eIFP-MPC lowers oscillations to 14.25 and produces a slightly shorter path (78.70\,cm vs. 80.79\,cm), though it incurs a higher completion time of 502.26\,s. These outcomes, illustrated in Fig.~\ref{fig:experiments}(f)--(g) and detailed in Table~\ref{tab:eifp_mpc_grouped_flat}, underscore the reduced instability achieved by eIFP-MPC, especially in scenarios involving directional opposition and high interaction density.

\subsection{Temporal Evolution of Trajectories}
Fig.~\ref{fig:eifpc_mpc} illustrates how the global plans evolve over time in a single four-robot scenario. Early stages (a)--(b) show some erratic shifts as robots adapt to each other’s changing velocities---for example, the red robot initially plans left in (a), then switches right in (b) after the blue robot commits to its right side. As interactions stabilize, the global plans in (c)--(e) become smoother and more consistent, reflecting eIFP-MPC’s ability to adapt early and converge to stable, coordinated paths.

\subsection{Scalability Analysis}
The scalability of the proposed planning approaches is examined in high-density multi-robot scenarios, specifically Setups F and G, which involve 12 and 16 robots, respectively. As the number of agents increases, traditional methods such as IFP encounter significant performance degradation due to compounded oscillations and planning failures. Although direct IFP results are not included for these dense settings, its inability to resolve symmetry and congestion---already evident in smaller configurations---would only worsen with scale. In contrast, both eIFP and eIFP-MPC remain effective, with eIFP-MPC consistently demonstrating lower oscillation rates and smoother trajectories under load. For instance, in Setup F (12 robots), eIFP records 86.08 oscillations and a time to goal of 90.66 seconds, while eIFP-MPC significantly reduces oscillations to 9.67 and slightly shortens the path length (80.11 cm vs. 81.77 cm), albeit at a longer goal time of 197.50 seconds due to MPC’s dynamic constraints. Similar performance gains are evident in Setup G (16 robots), where eIFP yields 25 oscillations and 336.45 seconds to goal, compared to eIFP-MPC’s reduced 14.25 oscillations and improved path efficiency (78.70 cm), with a trade-off in duration (502.26 seconds). These quantitative comparisons underscore the robustness of the proposed methods under increasingly congested, real-time conditions.

These results are further supported by the metric plots in Fig.~\ref{fig:plots} and the trajectory behaviors shown in Fig.~\ref{fig:experiments}(f)--(g). The plots reveal that while eIFP exhibits faster planning times in high-density environments---such as 44.78 ms for 16 robots versus 67.72 ms for eIFP-MPC---this comes at the cost of erratic motion and longer, less stable paths. On the other hand, eIFP-MPC's slightly longer planning time yields substantial benefits in path consistency, dynamic feasibility, and oscillation mitigation. Even as agent density and directional flow complexity increase, eIFP-MPC maintains smooth, stable motion patterns across most agents.

This trade-off highlights the algorithm’s practical applicability in large-scale deployments, where planning reliability outweighs marginal increases in computational time.

\section{Conclusion and Future Work}

This work presented enhancements to the Iterative Forecast Planner (IFP) aimed at improving motion planning in dense and symmetric multi-robot environments. By introducing time-to-collision (TCPA)-based obstacle prioritization, a cost-aware via-point selection mechanism with ghost-goal projection, and integrating a nonlinear Model Predictive Controller (MPC) for dynamic feasibility, the proposed framework—eIFP and its MPC-augmented variant—demonstrated robust, collision-free behavior in decentralized settings without communication. Extensive simulations across a range of scenarios revealed significant reductions in oscillations, smoother path execution, and improved performance under increasing robot densities. Future work will explore extensions to 3D navigation, support for heterogeneous agents, and deployment on real-world robotic platforms under sensor uncertainty, latency, and actuation constraints.


\begin{thebibliography}{00}
\bibitem{b1} J. Queralta et al., “Collaborative Multi-Robot Search and Rescue: Planning, Coordination, Perception, and Active Vision,” IEEE Access, vol. 8, pp. 191617–191643, 2020. doi: 10.1109/ACCESS.2020.3030190.
\bibitem{b2} A. Wang, “Intelligent warehouse multi-robot scheduling system based on improved A* algorithm,” in 2024 3rd Conf. Fully Actuated System Theory Appl. (FASTA), Shenzhen, China, 2024, pp. 1305–1310. doi: 10.1109/FASTA61401.2024.10595280.
\bibitem{b3} X. Yao, J. Zhang, and J. Oh, “Following social groups: Socially compliant autonomous navigation in dense crowds,” arXiv preprint, arXiv:1911.12063, 2019.
\bibitem{b4} X. Lin, Y. Huang, F. Chen, and B. Englot, “Decentralized multi-robot navigation for autonomous surface vehicles with distributional reinforcement learning,” in 2024 IEEE Int. Conf. Robotics Autom. (ICRA), 2024, pp. 8327–8333.
\bibitem{b5} C. Chakraa, E. Leclercq, F. Guérin, and D. Lefebvre, “Integrating collision avoidance strategies into multi-robot task allocation for inspection,” Trans. Inst. Meas. Control, vol. 47, no. 7, pp. 1466–1477, 2025.
\bibitem{b6} I. Solis, J. Motes, R. Sandström and N. M. Amato, "Representation-Optimal Multi-Robot Motion Planning Using Conflict-Based Search," in IEEE Robotics and Automation Letters, vol. 6, no. 3, pp. 4608-4615, July 2021.
\bibitem{b7} D. Matos, P. Costa, H. Sobreira, et al., “Efficient multi-robot path planning in real environments: a centralized coordination system,” Int. J. Intell. Robot. Appl., vol. 9, pp. 217–244, 2025. doi: 10.1007/s41315-024-00378-3.
\bibitem{b8} Y. Bai, S. Kotpalliwar, C. Kanellakis, et al., “Multi-agent Path Planning Based on Conflict-Based Search (CBS) Variations for Heterogeneous Robots,” J. Intell. Robot Syst., vol. 111, p. 26, 2025. doi: 10.1007/s10846-025-02229-0.
\bibitem{b9} B. Asfora, J. Banfi, and M. Campbell, “Mixed-Integer Linear Programming Models for Multi-Robot Non-Adversarial Search,” IEEE Robot. Autom. Lett., vol. 5, no. 4, pp. 6805–6812, Oct. 2020.
\bibitem{b10} K. Shi, L. Yang, Z. Wu, B. Jiang, and Q. Gao, “Multi-robot dynamic path planning with priority based on simulated annealing,” J. Franklin Inst., vol. 362, no. 1, 2025.
\bibitem{b11} N. E. Hwang, H. J. Kim, and J. G. Kim, “Centralized Mission Planning for Multiple Robots Minimizing Total Mission Completion Time,” Appl. Sci., vol. 13, no. 6, p. 3737, 2023. doi: 10.3390/app13063737. 
\bibitem{b12} V. Tuck, Y. V. Pant, S. A. Seshia, and S. S. Sastry, “DEC-LOS-RRT: Decentralized Path Planning for Multi-robot Systems with Line-of-sight Constrained Communication,” in 2021 IEEE Conf. Control Technol. Appl. (CCTA), San Diego, CA, USA, 2021.
\bibitem{b13} B. Şenbaşlar, W. Hönig, and N. Ayanian, “RLSS: real-time, decentralized, cooperative, networkless multi-robot trajectory planning using linear spatial separations,” Auton. Robot., vol. 47, pp. 921–946, 2023. doi: 10.1007/s10514-023-10104-w.
\bibitem{b14} Y. Lyu, J. M. Dolan, and W. Luo, “Decentralized Safe Navigation for Multi-agent Systems via Risk-aware Weighted Buffered Voronoi Cells,” in Proc. 2023 Int. Conf. Autonomous Agents Multiagent Syst. (AAMAS '23), Richland, SC, 2023, pp. 1476–1484.
\bibitem{b15} M. Everett, Y. F. Chen and J. P. How, "Motion Planning Among Dynamic, Decision-Making Agents with Deep Reinforcement Learning," 2018 IEEE/RSJ International Conference on Intelligent Robots and Systems (IROS), Madrid, Spain, 2018.
\bibitem{b16} H. Zhu, B. Brito, and J. Alonso-Mora, “Decentralized probabilistic multi-robot collision avoidance using buffered uncertainty-aware Voronoi cells,” Auton. Robots, vol. 46, no. 2, pp. 401–420, 2022.
\bibitem{b17} C. Wen et al., “Socially-Aware Robot Navigation Enhanced by Bidirectional Natural Language Conversations Using Large Language Models,” arXiv preprint, arXiv:2409.04965, 2024.
\bibitem{b18} S. Bakker, L. Knoedler, M. Spahn, W. Böhmer and J. Alonso-Mora, "Multi-Robot Local Motion Planning Using Dynamic Optimization Fabrics," 2023 International Symposium on Multi-Robot and Multi-Agent Systems (MRS), Boston, MA, USA, 2023.
\bibitem{b19} J. Ma, G. Chen, P. Jiang, Z. Zhang, J. Cao, and J. Zhang, “Distributed multi-robot obstacle avoidance via logarithmic map-based deep reinforcement learning,” in Proc. Third Int. Conf. Artif. Intell. Comput. Eng. (ICAICE 2022), vol. 12610, pp. 65–72, SPIE, Apr. 2023.
\bibitem{b20} H. Hailu, A. Yorozu, and A. Ohya, “Experimental Investigation of Mutual Avoidance Behavior for Multiple Autonomous Robots,” in Proc. 38th Japan Robot. Soc. Conf. (CD-ROM), Tsukuba, Japan, 2020, p. ROMBUNNO.1I2-04.
\bibitem{b21} A. Yorozu, H. Hailu and A. Ohya, "Experimental Investigation of Mutual Collision Avoidance Behavior for Multiple Mobile Robots," 2021 IEEE/ASME International Conference on Advanced Intelligent Mechatronics (AIM), Delft, Netherlands, 2021.
\bibitem{b22} J. Wang, Z. Yu, D. Zhou, J. Shi, and R. Deng, “Vision-based deep reinforcement learning of UAV autonomous navigation using privileged information,” arXiv preprint, arXiv:2412.06313, 2024.
\bibitem{b23} L. Chen, Y. Zhao, H. Zhao, and B. Zheng, “Non-Communication Decentralized Multi-Robot Collision Avoidance in Grid Map Workspace with Double Deep Q-Network,” Sensors, vol. 21, no. 3, p. 841, 2021. doi: 10.3390/s21030841.
\bibitem{b24} Z. Hu et al., “Bi-cl: A reinforcement learning framework for robots coordination through bi-level optimization,” in 2024 IEEE/RSJ Int. Conf. Intell. Robots Syst. (IROS), 2024.
\bibitem{b25} B. Şenbaşlar and G. S. Sukhatme, "DREAM: Decentralized Real-Time Asynchronous Probabilistic Trajectory Planning for Collision-Free Multirobot Navigation in Cluttered Environments," in IEEE Transactions on Robotics, vol. 41, pp. 573-592, 2025.
\bibitem{b26} M. Jankovic, M. Santillo, and Y. Wang, “Multiagent Systems With CBF-Based Controllers: Collision Avoidance and Liveness From Instability,” IEEE Trans. Control Syst. Technol., vol. 32, no. 2, pp. 705–712, 2024.
\bibitem{b27} K. Zhang, M. Zahmatkesh, M. Stefanec, F. Arvin and J. Hu, "A Real-Time RRT-APF Approach for Efficient Multi-Robot Navigation in Complex Environments," 2025 IEEE International Conference on Industrial Technology (ICIT), Wuhan, China, 2025.
\bibitem{b28} L. Pan, K. Hsu and N. Ayanian, "Hierarchical Large Scale Multirobot Path (Re)Planning," 2024 IEEE/RSJ International Conference on Intelligent Robots and Systems (IROS), Abu Dhabi, United Arab Emirates, 2024.
\bibitem{b29} J. Pushpangathan and H. Kandath, “RRT and velocity obstacles-based motion planning for unmanned aircraft systems traffic management (UTM),” arXiv preprint, arXiv:2302.14543, 2023.
\bibitem{b30} Z. Lu, K. Feng, J. Xu, H. Chen, and Y. Lou, “Robot safe planning in dynamic environments based on model predictive control using control barrier function,” arXiv preprint, arXiv:2404.05952, 2024.
\end{thebibliography}
\end{document}